\documentclass[10pt,twocolumn,letterpaper]{article}

\usepackage{3dv}
\usepackage{times}
\usepackage{epsfig}
\usepackage{graphicx}
\usepackage{amsmath}
\usepackage{amssymb}
\usepackage[utf8]{inputenc}
\usepackage[symbol]{footmisc}
\DefineFNsymbols{mySymbols}{{\ensuremath\dagger}{\ensuremath\ddagger}\S\P
   *{**}{\ensuremath{\dagger\dagger}}{\ensuremath{\ddagger\ddagger}}}
\setfnsymbol{mySymbols}
\usepackage{booktabs}

\usepackage[pagebackref=true,breaklinks=true,letterpaper=true,colorlinks,bookmarks=false]{hyperref}

\threedvfinalcopy 

\def\threedvPaperID{128} 

\ifthreedvfinal\fi
\begin{document}

\title{Learning to Estimate Indoor Lighting from 3D Objects}

\author{Henrique Weber$^1$, Donald Prévost$^2$, and Jean-François Lalonde$^{1,2}$\thanks{Contact author: \tt\small jflalonde@gel.ulaval.ca}\\
$^1$Université Laval, $^2$Institut National d'Optique
}

\maketitle
\thispagestyle{empty}

\begin{abstract}
   In this work, we propose a step towards a more accurate prediction of the environment light given a single picture of a known object. To achieve this, we developed a deep learning method that is able to encode the latent space of indoor lighting using few parameters and that is trained on a database of environment maps. This latent space is then used to generate predictions of the light that are both more realistic and accurate than previous methods. To achieve this, our first contribution is a deep autoencoder which is capable of learning the feature space that compactly models lighting. Our second contribution is a convolutional neural network that predicts the light from a single image of an object with known geometry and reflectance. To train these networks, our third contribution is a novel dataset that contains 21,000 HDR indoor environment maps. The results indicate that the predictor can generate plausible lighting estimations even from diffuse objects.
\end{abstract}

\section{Introduction}

Estimating lighting from images enables a wide range of possible applications, ranging from the realistic insertion of virtual content in augmented reality~\cite{gardner-sigasia-17,mandl2017learning,karsch2014automatic}, shadow or highlight removal~\cite{jiddi2016reflectance}, image matching~\cite{yu2012novel}, appearance transfer~\cite{liu2017material} or reflectance and/or geometry estimation~\cite{lombardi2016radiometric} to name just a few. However, estimating light from an image is a challenging problem. Indeed, light is just one of the many components in the image formation process, which also involves camera parameters, scene material reflectance, the geometry of objects and other post-processing operations. Disentangling one of these components from the others is an ill-posed inverse problem, since infinitely many of their combinations can create the same image. 

In the literature, this problem is typically solved with two different alternatives: 1) placing a specially-designed object (a light probe) into the scene (either reflective~\cite{Debevec-siggraph-98,debevec-sslp-12} or not~\cite{calian-sigtr-13}); or 2) imposing constraints on the models used to represent lighting (e.g. low frequency spherical harmonics, or SH~\cite{green2003spherical}). Unfortunately, both of these solutions come with significant disadvantages. Indeed, it is often not possible or practical to insert a light probe in the scene. In addition, the large variety of light sources indoors (ranging from high frequency halogen lights to large area lights such as windows) make it difficult to find a low-dimensional lighting representation which will match them all well. 

In this work, we propose a solution to both of these issues. First, we rely on a common object already present in the scene to estimate the lighting instead of using specially designed light probes. While the pose, geometry and reflectance of the object must be known (existing object detection~\cite{hinterstoisser2012model} and/or tracking~\cite{garon-tvcg-17} methods can be used to obtain this information), the object itself needs not to be specially-designed for the purpose of lighting estimation~\cite{calian-sigtr-13} and can be any common object. Second, instead of restraining ourselves to low-frequency lighting models, we propose to learn the space of indoor lighting environments. For this, we train a deep autoencoder on a novel dataset of HDR indoor environment maps~\cite{gardner-sigasia-17}. The autoencoder learns to compress lighting environments to a low-dimensional latent vector. We then train another convolutional neural network which learns to map the image of a particular object to that latent space. We demonstrate that our approach is more accurate than estimating SH lighting from the full transport matrix. Being much faster, our approach is also more amenable to real-time augmented reality scenarios. 

In short, we make the following key contributions: 
\begin{itemize}
	\item A new method for robustly estimating lighting from a known object which outperforms the state-of-the-art;
	\item A thorough evaluation on a large dataset of synthetic objects, comparing to previous approaches; 
	\item A demonstration of the applicability of our method on real data; 
	\item A novel dataset of 21,000 indoor lighting environments that, along with the code, will be released to the community. 
\end{itemize}



\section{Related work}

Estimating lighting from images, or inverse lighting, has a long history in computer vision. Some techniques rely on capturing mirrored and diffuse spheres~\cite{Debevec-siggraph-98,reinhard-book-05,debevec-sslp-12} to directly recover omnidirectional HDR lighting. Follow-up work has also proposed to design objects specifically designed to capture and reproduce shading in real-time~\cite{calian-sigtr-13}. Unfortunately, relying on specific objects can limit the applicability of the approach, so while our technique also relies on known objects, it does not impose a constraint on the type of object used. 

A large number of techniques estimate lighting directly from images. One of the earlier works~\cite{marschner1997inverse} estimate lighting as a sum of basis functions on the sphere. They used regularization during the regression process to deal with the ill-conditioned nature of the problem. Instead of explicitly regularizing the estimation, Ramamoorthi and Hanrahan~\cite{ramamoorthi2001efficient} propose to use spherical harmonics with a small number of coefficients to represent light. Thus, the parametric model itself serves as a regularizer to constrain the space of solutions. In a subsequent paper~\cite{ramamoorthi2001signal}, the same authors frame the estimation of spherical harmonics coefficients as a linear least squares problem, which we will use in our experiments below. Approaches that robustly solve for illumination and reflectance~\cite{lombardi2016reflectance}, as well as geometry~\cite{barron2015shape} must rely on strong learned priors to help constrain the optimization. 

Due to their ubiquity in photographs, human faces are often used to estimate lighting. Typically, inverse lighting is performed to improve other face-related tasks, such as face recognition~\cite{wen-cvpr-03}, face geometry~\cite{kemelmacher-pami-11} or texture~\cite{li-eccv-14} recovery, or face swapping~\cite{bitouk-siggraph-08} in graphics. Of particular relevance to our work, Calian et al.~\cite{calian-eurographics-18} recently proposed estimate HDR outdoor lighting from a face image using inverse rendering, but with the particularity that they learn a low-dimensional space for outdoor lighting with a deep autoencoder. Here, we apply similar ideas to the context of indoor lighting, and to more generic objects. 

Augmented reality applications also benefit from robust lighting estimates. In this context, real-time performance is critical. Gruber et al.~\cite{gruber2012real} model the Radiance Transfer Function of an entire scene from its captured geometry, but it requires from 50 to 200 ms per frame. To reduce the time complexity, and bearing similarities to our approach, Mandl et al.~\cite{mandl2017learning} uses deep learning to estimate the SH coefficients from an object. In our case, we do not learn SH coefficients but instead learn to map an image to the latent space of indoor lighting. 

Finally, deep learning has recently been used in many contexts for lighting estimation as well. For example, Gardner et al.~\cite{gardner-sigasia-17} estimate HDR illumination from a single indoor image, while Hold-Goeffroy et al.~\cite{holdgeoffroy-cvpr-17} do so outdoors by relying on a physically-based sky model. In \cite{rematas2016deep}, the reflectance map of an object (i.e., its ``orientation-dependent'' appearance) is estimated from a single image using a CNN. Follow-up work~\cite{georgoulis2016delight} learns to separate a reflectance map into material and illumination estimation with two different CNNs and \cite{Georgoulis_2017_ICCV} show lighting estimation from multiple (specular) materials. In contrast, our approach assumes knowledge of the object, but does not depend on the presence of multiple specular materials. 

\section{Method}

\begin{figure*}
\centering
\includegraphics[width=1\linewidth]{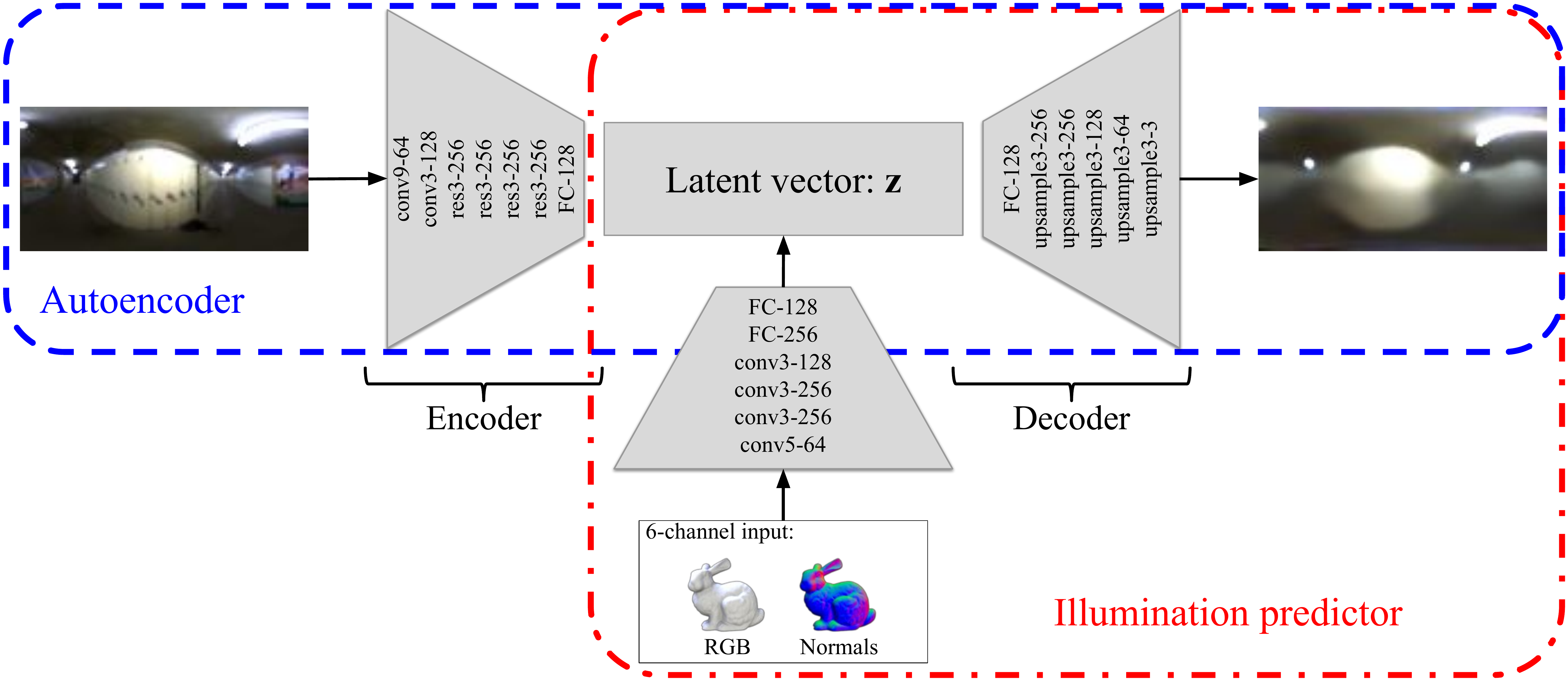}
\caption{Our proposed network for recovering the environment map from a single image. The notation ‘conv\textit{x-y}’ means a convolution layer with \textit{x} dimensions and \textit{y} filters, ‘res\textit{x-y}’ is a residual block with \textit{x} dimensions and \textit{y} filters, ‘upsample\textit{x-y}’ is a convolution followed by upsampling, while ‘FC-\textit{x}’ represents a fully-connected layer of \textit{x} units. The encoder compresses the input $\mathbf{e}$ to a $128$-dimensional vector $\mathbf{z}$ through $4$ convolutional layers. $\mathbf{z}$ is later decompressed by $4$ deconvolutional filters at the decoder to form the output image $\mathbf{\hat{e}}_\text{AE}$. Next, we have the illumination predictor. It receives as input an RGB image and the normal map of the object to perform a series of convolutions to output a 128-dimensional vector $\mathbf{\hat{z}}$. This vector is then sent to the decoder which outputs an environment map $\mathbf{\hat{e}}_\text{i}$. At test time, we just use the the illumination predictor and the decoder.}
\label{fig:network}
\end{figure*}

Our proposed method is divided in two parts. The first part learns to represent the space of indoor lighting in a compact way with an autoencoder, which we train on a large dataset of indoor lighting conditions. The second part learns to map the image of an object to the latent space of the autoencoder. In that way, our method bears resemblance to the ``T-network'' architecture of~\cite{girdhar2016learning}, but is targeted towards learning illumination. To train the two neural networks, we need a set of $N$ image pairs, each containing: an HDR environment map $\mathbf{e}_i$, $i\in\{1,...,N\}$ and the corresponding image $\mathbf{i}_i$ of the object of interest lit by $\mathbf{e}_i$. Sec.~\ref{sec:dataset} will describe how we obtain such a dataset, but let us assume for now that it is provided.



\subsection{Learning the space of indoor lighting}
\label{sec:learning-ae}

We choose to represent a lighting environment with a non-parametric model trained on a dataset of indoor HDR panoramas. This is done with a convolutional autoencoder, which is capable of learning a representation (encode) from training data. It is also able to generate data (decode) from a given sample that lies on this learned feature space.

The autoencoder (see fig.~\ref{fig:network}) takes as input an RGB HDR panorama $\mathbf{e}$ of size $64 \times 128$ pixels represented in the equirectangular format. It is then passed through $2$ convolutional layers, $4$ residual blocks~\cite{he2016deep} and a fully-connected layer until it is compressed to a $Z$-dimensional latent vector $\mathbf{z}$. This set of operations is denoted by
\begin{equation}
\mathbf{z} = f_\text{enc}(\mathbf{e}) \,,
\label{eq:encoder}
\end{equation}
where $f_\text{enc}(\cdot)$ is the \emph{encoder} (see fig.~\ref{fig:network}). The vector $\mathbf{z}$ is decompressed by another fully connected layer and $4$ upsampling layers to output an approximate version of the input. This second set of operations is denoted by
\begin{equation}
\mathbf{\hat{e}} = f_\text{dec}(\mathbf{z}) \,,
\label{eq:decoder}
\end{equation}
where $\mathbf{\hat{e}}$ is the reconstructed environment map from the latent vector $\mathbf{z}$, and $f_\text{dec}(\cdot)$ is the \emph{decoder} (see fig.~\ref{fig:network}). Batch normalization and the ELU activation function ~\cite{clevert2015fast} are used in all layers, except in the output layer of the encoder and decoder.

We train the autoencoder to minimize the solid angle-weighted L1-loss over all $N$ environment maps in a training dataset: 
\begin{equation}
\mathcal{L}_\text{AE} = \sum_{i=1}^N || \mathbf{w} \odot \left( \log(\mathbf{e}_i+1) - \log(f_\text{dec}(f_\text{enc}(\mathbf{e}_i))+1) \right) ||_1 \,,
\label{eq:loss-ae}
\end{equation}
where $\mathbf{w}$ is a matrix containing the solid angle of each pixel and $\odot$ denotes pixel-wise multiplication. Here, the log is used to compress the potentially very high dynamic range of indoor lighting into more manageable values. 


\subsection{Learning to estimate lighting from an object}
\label{sec:learning-ip}

Once the autoencoder has learned a latent representation for indoor lighting, we train another neural network whose task is to map the appearance of an object to that latent space (see the bottom part of fig.~\ref{fig:network}). This network takes as input a $128 \times 128$ LDR image of an object as well as a normal map of the same object concatenated as a 6-channel input $\mathbf{i}$. It is then passed through a series of $4$ convolutional and $2$ fully-connected layers that bring the input to the $Z$-dimensional latent vector $\mathbf{z}$. This vector is then decompressed by the (already trained) $f_\text{dec}$ of the first network, which generates an estimation of the light in the scene: 
\begin{equation}
\mathbf{\hat{z}} = f_\text{ip}(\mathbf{i}) \,,
\label{eq:image-predictor}
\end{equation}
where $\mathbf{\hat{z}}$ is the latent vector estimated from the input image $\mathbf{i}$ by the illumination predictor $f_\text{ip}(\cdot)$. This illumination predictor is trained to minimize the L2 loss between its prediction $f_\text{ip}(\mathbf{i}_1)$ given the object image and the latent vector of the environment map used to light the object $f_\text{enc}(\mathbf{e}_i)$: 
\begin{equation}
\mathcal{L}_\text{IP} = \sum_{i=1}^N || f_\text{enc}(\mathbf{e}_i) - f_\text{ip}(\mathbf{i}_i) ||_2 \,.
\label{eq:loss-ip}
\end{equation}
%

\section{Datasets}
\label{sec:dataset}

We rely on two datasets to both train and quantify the results of our proposed method. One is a dataset of HDR indoor panoramas that allowed us to train both the autoencoder and the illumination predictor. The other dataset is composed of renders of different objects, which is used to train and quantify the accuracy of the illumination predictor network.

\subsection{HDR indoor panoramas}
\label{sec:dataset_indoor_panoramas}

Although there exists some HDR datasets in the literature, they are either small with tens of panoramas like hdrLabs~\cite{hdrlabs}, contain saturated HDR (Matterport database~\cite{Matterport3D}) or are outdoors only (Laval HDR Sky Database~\cite{hdrdb,lalonde-3dv-14}). Therefore, we introduce a novel dataset that contains $21,000$ HDR indoor environment maps. To do so, we rely on the Laval Indoor HDR panorama dataset~\cite{gardner-sigasia-17}, which is composed of $2,300$ $360^\circ$ panoramas shot in full HDR in a variety of indoor scenes ranging from kitchens and basements to churches and grocery floors. We begin by selecting a subset of the panoramas, trying to cover the greatest variability possible. Also, since some panoramas in this dataset were shot in similar scenes (like two panoramas shot in the same kitchen of the same house), we avoid selecting pictures that were shot in less than 10 minutes before or after any other panorama that was already selected. After this screening, we ended up with 1,303 images, and split them into training (1,044), validation (159) and testing (100).

Next, the main surfaces in each panorama were labeled with the scribble-based interface of EnvyDepth~\cite{banterle2013envydepth}, which then assigns depth values to each pixel in the environment map according to the label. The final output is then a set of virtual point lights (VPLs) (one for each scene surface) represented by position, normal, color, and a scale to preserve the total energy of the environment map given as input. 

\begin{figure}[t]
\centering
\includegraphics[width=\linewidth]{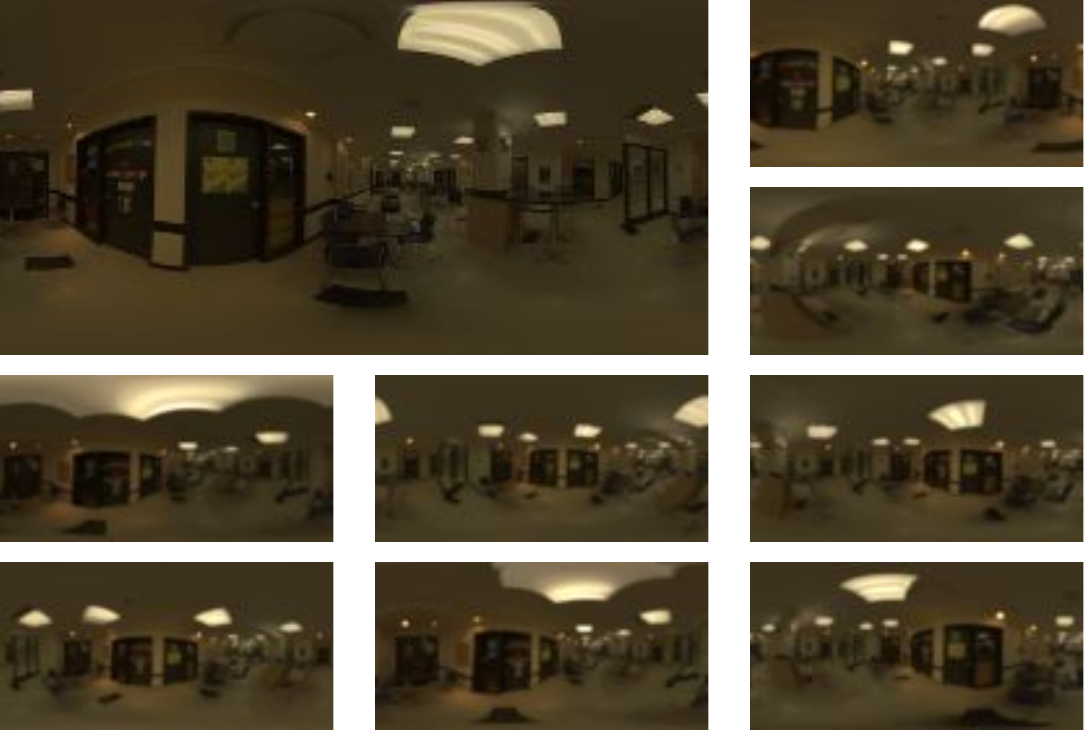}
\caption{The top-left panorama was taken from the Laval Indoor HDR Dataset~\cite{gardner-sigasia-17}. By using our data-augmentation approach, we can warp this panorama in a geometrically-consistent way, which effectively generates new lighting environment maps such as the ones shown around the original panorama. In particular, one can see the light on the roof with a great distortion among distinct renders.}
\label{fig:envydepth}
\end{figure}

With this extra information in hand, we can now augment the dataset by rendering multiple versions of each panorama. For that, we take all the VPLs from a given environment map and sample a random camera pose that remains inside the VPLs perimeter (under the ceiling, over the floor and between the walls of the modeled scene). Since the point cloud formed by the VPLs is sparse, we insert a diffuse sphere that surrounds the scene to prevent black holes in the rendered panorama. The physically-based rendering engine Mitsuba~\cite{mitsuba} is then used to render a new latitude-longitude panorama of $64 \times 128$ pixel resolution in the equirectangular format, by placing a virtual omnidirectional camera at the sampled camera pose. Examples of panoramas created with this approach are shown in fig.~\ref{fig:envydepth}. Note how the resulting panoramas are both geometrically plausible yet quite different from one another, even if they all originate from the same one. 

\subsection{LDR renders of the target object}
\label{sec:dataset_ldr_renders}

To train the illumination estimation network, we rely on synthetic images of the object of interest lit by different lighting conditions. The background is masked out, and the outputs consists of the RGB image and a normal map. We make use of five 3D models: bunny, dragon, buddha, sphere and spiky sphere. Next, each model is rendered with three distinct materials: diffuse (albedo of $0.5$), rough plastic (interior and exterior index of refraction of $1.9$ and $1$, respectively, and diffuse albedo of $0.5$), and conductor (with measured copper data). 

\begin{figure}[t]
\centering
\includegraphics[width=\linewidth]{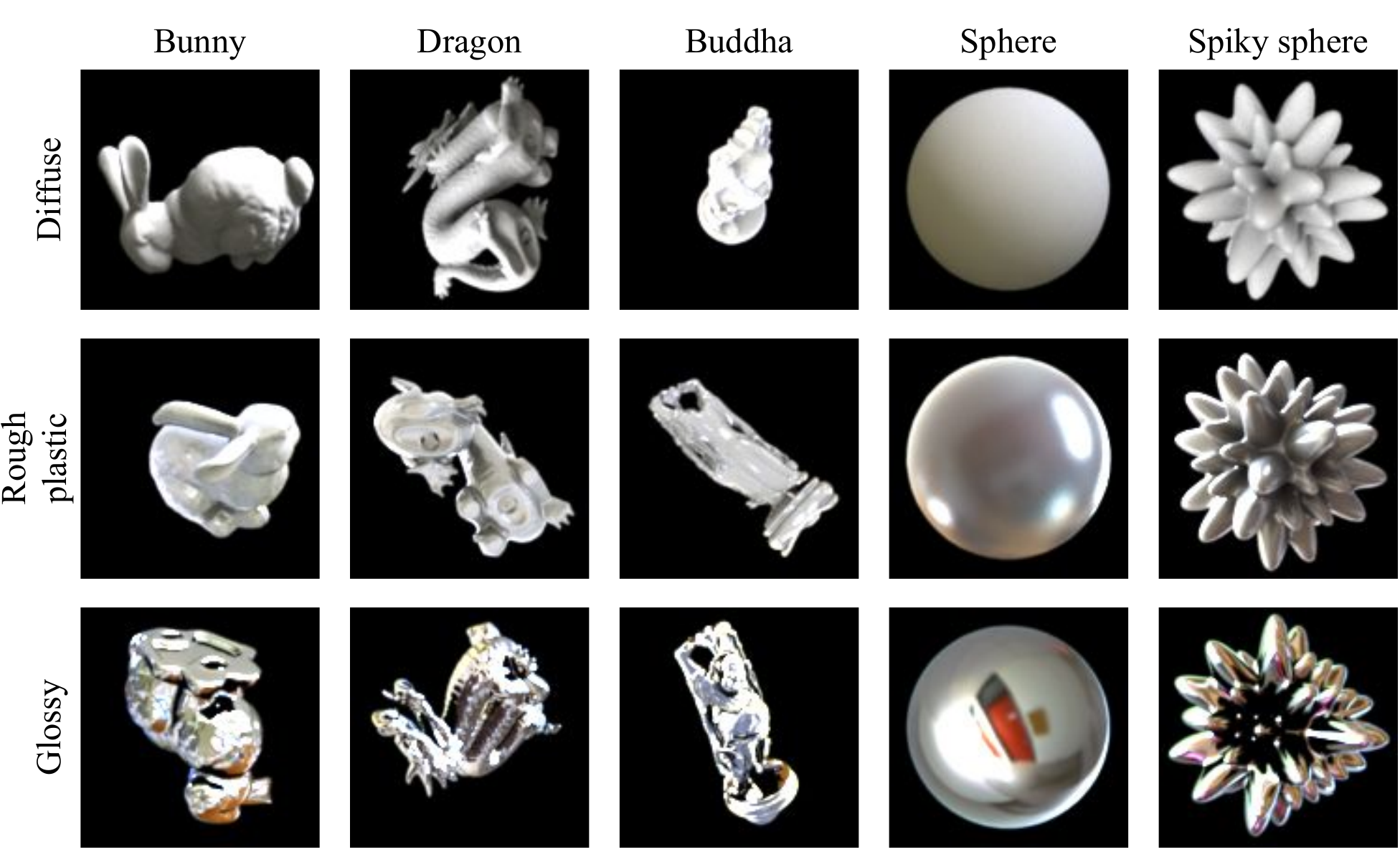}
\caption{Examples of renders from our dataset of synthetic data. Each row represents objects with diffuse, rough plastic, and conductor materials, respectively.}
\label{fig:ldr_dataset}
\end{figure}

To render the object, the model is first placed at the origin of the world coordinate system, and is rotated randomly. To calculate this random pose in spherical coordinates ($\theta,\phi$), we first select a random $\theta \in [-180^\circ, 180^\circ)$ and then set $\phi= \cos^{-1}(2x - 1)$ (in radians) with $x \in [0,1)$. A random environment map is obtained from the dataset, and also gets rotated randomly with the same sampling procedure. For each object, approximately 30K images are generated for training, and 5K for validation (using the same split of panoramas as the one explained in sec.~\ref{sec:dataset_indoor_panoramas}). As before, the Mitsuba~\cite{mitsuba} rendering engine is used to generate all renders. The resulting images are then re-exposed (by mapping the 90th percentile to 0.8) and clipped in the $[0,1]$ interval to generate linear LDR images. Examples of renders using this approach can be seen in fig.~\ref{fig:ldr_dataset}.

\section{Experiments}
\label{sec:experiments}

\begin{figure*}[t]
\centering
\includegraphics[width=1\linewidth]{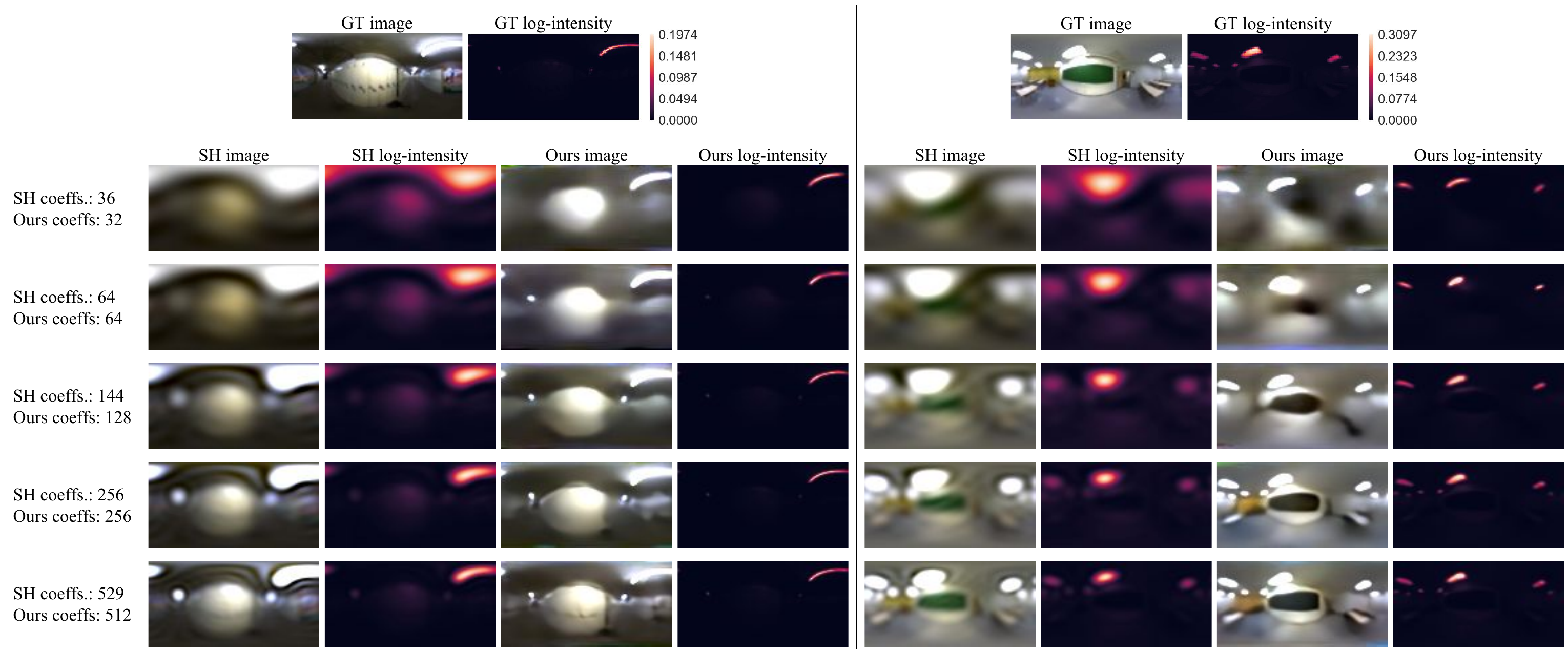}
\caption{Qualitative comparison for direct fits with spherical harmonics of degree $\{5, 7, 11, 15, 22\}$. The colored version of the environment maps are tone mapped with $\gamma=1.4$ for display, while the heat map shows the log-intensity of the estimations clipped at the 95th percentile of the GT intensity. Also for display purposes, the SH estimations are low-pass filtered to avoid ringing artifacts.}
\label{fig:qualitative_autoencoder}
\end{figure*}

\begin{table*}[]
\centering
\begin{tabular}{llllllllll}
\toprule
\multicolumn{2}{l}{\textbf{\begin{tabular}[c]{@{}l@{}}Number of \\ coefficients\end{tabular}}} & \multicolumn{2}{l}{\textbf{si-RMSE}} & \multicolumn{2}{l}{\textbf{RMSE}} & \multicolumn{2}{l}{\textbf{MAE}} & \multicolumn{2}{l}{\textbf{MRE}} \\
\midrule
SH & Ours & SH & Ours & SH & Ours & SH & Ours & SH & Ours \\ 
\midrule
36 & 32 & 0.0602 & \textbf{0.0338} & 0.0602 & \textbf{0.0432} & 0.0163 & \textbf{0.0063} & 4.0838 & \textbf{1.2092} \\
64 & 64 & 0.0537 & \textbf{0.0310} & 0.0537 & \textbf{0.0356} & 0.0150 & \textbf{0.0048} & 3.8326 & \textbf{1.0216} \\
144 & 128 & 0.0437 & \textbf{0.0220} & 0.0437 & \textbf{0.0287} & 0.0133 & \textbf{0.0039} & 3.2325 & \textbf{0.7779} \\
256 & 256 & 0.0401 & \textbf{0.0181} & 0.0401 & \textbf{0.0206} & 0.0113 & \textbf{0.0030} & 2.8512 & \textbf{0.5828} \\
529 & 512 & 0.0305 & \textbf{0.0167} & 0.0306 & \textbf{0.0196} & 0.0088 & \textbf{0.0028} & 2.2758 & \textbf{0.5376} \\
\bottomrule
\end{tabular}%
\vspace{.5em}
\caption{Quantitative comparison between representing an environment map with our autoencoder with projecting it to the spherical harmonics basis with matching numbers of degrees of freedom. For the autoencoder, ``number of coefficients'' refers to $Z$, the dimension of the latent vector $\mathbf{z}$.}
\label{tab:direct_fit}
\end{table*}

We now proceed to evaluate the proposed method. The goal here is twofold: determine whether (1) the autoencoder is capable of compactly and accurately represent light; and (2) the illumination predictor can efficiently map the appearance of an object to the learned embedded space. To answer these questions, we calculate a variety of metrics that compare the ground-truth panorama $\mathbf{e}$ to the one obtained by the autoencoder network $\mathbf{\hat{e}}_\text{ae} = f_\text{dec}(f_\text{enc}(\mathbf{e}))$ (eqs.~(\ref{eq:encoder}) and (\ref{eq:decoder}), sec.~\ref{sec:learning-ae}) and the one generated by the illumination predictor $\mathbf{\hat{e}}_\text{ip} = f_\text{dec}(f_\text{ip}(\mathbf{i}))$ (eqs.~(\ref{eq:decoder}) and (\ref{eq:image-predictor}), sec.~\ref{sec:learning-ip}).

The evaluation is done over the test set of images described in sec.~\ref{sec:dataset}, which is composed by 100 panoramas. To quantify the errors, we use lighting-based metrics computed directly over the predicted and ground-truth panoramas. We compute the root mean squared error (RMSE), the scale-invariant RMSE (si-RMSE)~\cite{barron2015shape}, the mean absolute error (MAE) and the mean relative error (MRE, that is, the MAE divided by the ground truth value). All these metrics are weighted by solid angles. 

\subsection{Modeling indoor lighting}
\label{sec:indoor_light_modeling}

\begin{figure*}
\centering
\includegraphics[width=1\linewidth]{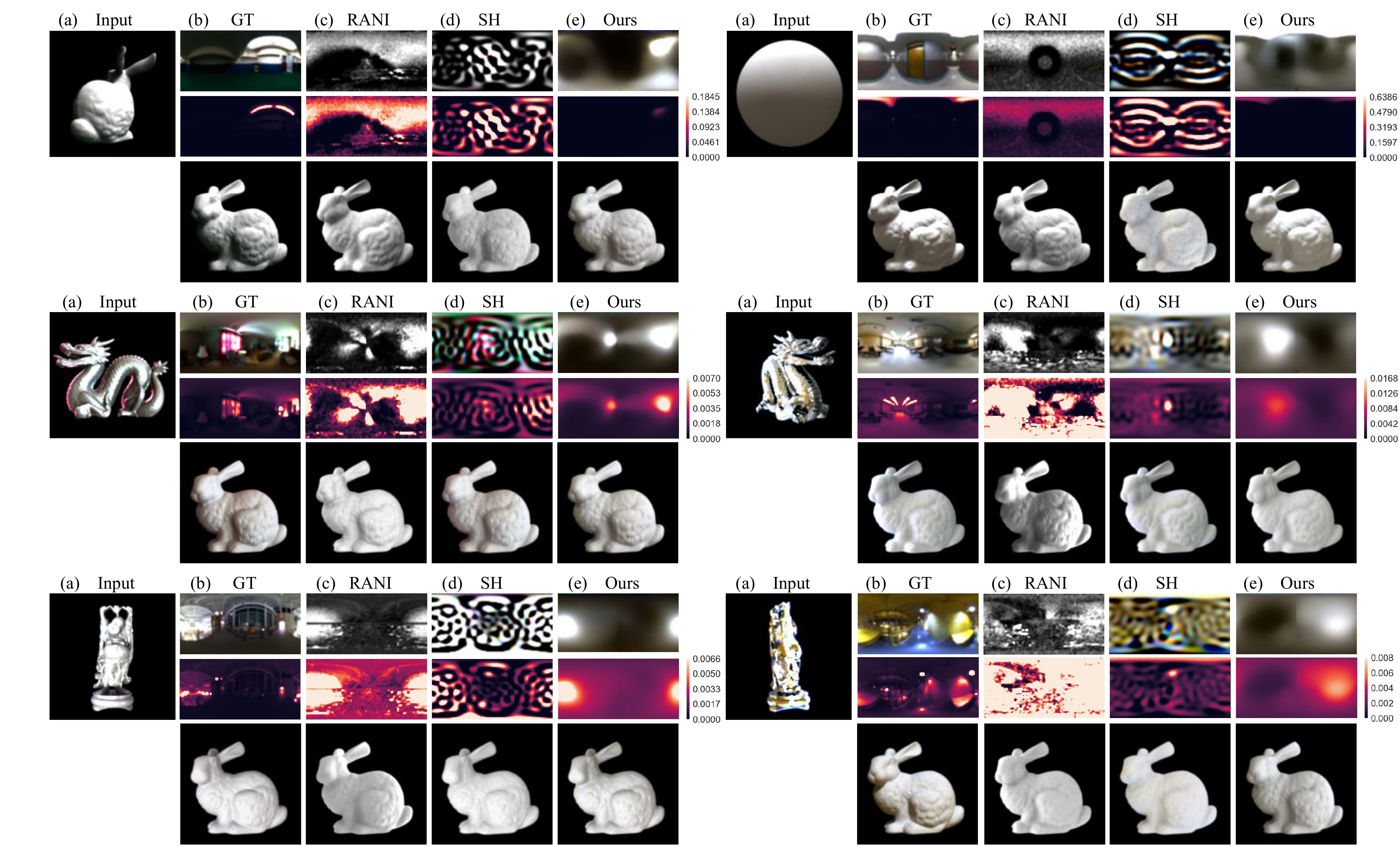}
\caption{Qualitative comparison of our approach on synthetic data with spherical harmonics~\cite{ramamoorthi2001signal} and RANI~\cite{lombardi2016reflectance}. Each result shows (a) the input image; (b) the ground truth lighting; and the results with (c) RANI~\cite{lombardi2016reflectance}, (d) SH~\cite{ramamoorthi2001signal}, and (e) our method. The last row shows a diffuse bunny relit with each light source.}
\label{fig:qualitative_illum_predict}
\end{figure*}

First, we present the accuracy of the autoencoder to represent indoor light directly. Quantitative results, comparing the performance of our autoencoder on all metrics with spherical harmonics with approximately the same number of degrees of freedom, are shown in table~\ref{tab:direct_fit}, while corresponding qualitative results are shown in fig.~\ref{fig:qualitative_autoencoder}. In particular, we experiment with $Z=\{32, 64, 128, 256, 512\}$ and corresponding SH coefficients $\{36, 64, 144, 256, 529\}$ (SH degree $\{5, 7, 11, 15, 22\}$ respectively). Overall, the autoencoder is capable of more accurately representing indoor lighting across all metrics, and all number of coefficients. Interestingly, the autoencoder is capable of representing the high-frequency light sources even with very few coefficients ($Z=32$). The quality of the reconstructed environment map significantly increases as $Z$ increases. 


\subsection{Lighting estimation from synthetic data}
\label{sec:lighting_estimation_synthetic}

We now proceed to evaluate the inverse lighting estimation from synthetic data. We compare our results against the ones obtained with the approach of \cite{ramamoorthi2001signal} (SH) and \cite{lombardi2016reflectance} (RANI). Since \cite{ramamoorthi2001signal} needs a pre-computed light transport matrix to recover spherical harmonics lighting, we consider this to be an upper performance bound for other techniques which also regress SH lighting (e.g.~\cite{barron2015shape,mandl2017learning}), and thus do not explicitly compare with those techniques. For these experiments, we rely on a single-bounce light transport matrix for each object and each pose. Here, we use $Z=128$. Even if more accurate lighting reconstructions can be obtained with higher values for $Z$ as seen in sec.~\ref{sec:indoor_light_modeling}, we experimentally determined that in the \emph{inverse} lighting case, higher values were not beneficial.

To generate the test dataset, we use the same 5 objects and 3 materials as in sec.~\ref{sec:dataset_ldr_renders}, but use a fixed set of 4 viewpoints: front, top, and side (object rotated by $120^\circ$ and $240^\circ$). Two viewpoints were selected for the spiky sphere while a single one was selected for the sphere. Note that since the viewpoints were generated randomly in the training set, it is extremely unlikely that those would overlap between the train and test set. The entire test set is thus composed of $3~\text{materials} \times (4 \times 3 + 2 + 1~\text{viewpoints}) \times 100~\text{envmaps} = 4,500$ renders. The metrics used here to evaluate the results are the same as the ones described above.

Table~\ref{tab:quantitative_synthetic} report quantitative comparisons between the approaches on the test dataset. Overall, our approach achieves the best results in the majority of the scenarios. In particular, we note how our approach achieves similar results across different materials. This is particularly interesting since it indicates that our approach can recover high frequency lighting even from low frequency (e.g. diffuse) material. When the objects are very glossy, then the transport matrix has high rank and can thus more reliably be inverted. Note that, as opposed to benefiting from the transport matrix, our approach relies only on the object surface normals as geometric input information. Qualitative results shown in fig.~\ref{fig:qualitative_illum_predict} illustrate visual examples for all 3 techniques. Our approach is much better at focusing on the most important light sources in the scene, while keeping the remaining part of the environment map smooth. 

\paragraph{Runtime comparison} RANI~\cite{lombardi2016reflectance} takes approximately 2 minutes per image on an Nvidia Titan X Pascal GPU, while SH~\cite{ramamoorthi2001signal} computes lighting in 0.55 seconds (once the transport matrix has been loaded in memory) on an Intel i7-6800K CPU at 3.40GHz. In contrast, our network estimates lighting in 6 ms per image on average on the same GPU used by ~\cite{lombardi2016reflectance}, making it suitable for real-time applications.

\begin{table*}[]
\centering
\resizebox{\textwidth}{!}{%
\begin{tabular}{llrrrrrrrlrrrlrrr}
\toprule
 &  & \multicolumn{3}{c}{\textbf{si-RMSE}} & \multicolumn{1}{c}{\textbf{}} & \multicolumn{3}{c}{\textbf{RMSE}} & \multicolumn{1}{c}{\textbf{}} & \multicolumn{3}{c}{\textbf{MAE}} & \multicolumn{1}{c}{\textbf{}} & \multicolumn{3}{c}{\textbf{MRE}} \\
\multicolumn{1}{r}{} & \multicolumn{1}{r}{} & RANI & SH & Ours &  & RANI & SH & Ours & \multicolumn{1}{r}{} & RANI & SH & Ours & \multicolumn{1}{r}{} & RANI & SH & Ours \\ 
\midrule
Bunny & Diffuse & 0.0667 & 0.0670 & \textbf{0.0632} &  & 0.1722 & 0.1967 & \textbf{0.0705} &  & 0.1029 & 0.0699 & \textbf{0.0129} &  & 47.2796 & 31.9370 & \textbf{2.9630} \\
 & Rough plastic & 0.0662 & 0.0660 & \textbf{0.0633} &  & 0.1122 & \textbf{0.0663} & 0.0705 &  & 0.0535 & 0.0148 & \textbf{0.0129} &  & 20.8638 & 3.8965 & \textbf{2.9703} \\
 & Glossy & 0.0666 & 0.0646 & \textbf{0.0645} &  & 0.1533 & \textbf{0.0653} & 0.0705 &  & 0.0680 & \textbf{0.0091} & 0.0127 &  & 29.1660 & \textbf{1.3840} & 2.9665 \\*[.5em]
 Dragon & Diffuse & \textbf{0.0636} & 0.0672 & 0.0651 &  & 0.1188 & 0.1975 & \textbf{0.0705} &  & 0.0521 & 0.0641 & \textbf{0.0130} &  & 29.2360 & 31.0396 & \textbf{2.9685} \\
 & Rough plastic & 0.0667 & 0.0655 & \textbf{0.0647} &  & 0.0997 & \textbf{0.0661} & 0.0705 &  & 0.0384 & 0.0182 & \textbf{0.0129} &  & 15.7579 & 5.2141 & \textbf{2.9717} \\
 & Glossy & 0.0666 & 0.0652 & \textbf{0.0648} &  & 0.1442 & \textbf{0.0654} & 0.0704 &  & 0.0661 & \textbf{0.0099} & 0.0129 &  & 23.9570 & \textbf{1.8808} & 2.9699 \\*[.5em]
Buddha & Diffuse & \textbf{0.0644} & 0.0695 & 0.0653 &  & 0.1416 & 0.1669 & \textbf{0.0705} &  & 0.0575 & 0.0500 & \textbf{0.0130} &  & 32.7316 & 32.4999 & \textbf{2.9723} \\
 & Rough plastic & \textbf{0.0645} & 0.0678 & 0.0665 &  & 0.1156 & 0.0725 & \textbf{0.0705} &  & 0.0429 & 0.0181 & \textbf{0.0130} &  & 17.0445 & 5.0871 & \textbf{2.9747} \\
 & Glossy & \textbf{0.0630} & 0.0666 & 0.0667 &  & 0.1412 & \textbf{0.0667} & 0.0705 &  & 0.0559 & \textbf{0.0109} & 0.0130 &  & 26.0428 & \textbf{2.4685} & 2.9741 \\*[.5em]
Sphere & Diffuse & 0.0663 & 0.0674 & \textbf{0.0472} &  & 0.1974 & 0.7465 & \textbf{0.0703} &  & 0.1400 & 0.3484 & \textbf{0.0129} &  & 59.4910 & 144.2276 & \textbf{2.9688} \\
 & Rough plastic & 0.0655 & 0.0667 & \textbf{0.0476} &  & 0.1276 & \textbf{0.0668} & 0.0703 &  & 0.0687 & \textbf{0.0127} & 0.0129 &  & 27.0281 & \textbf{2.7128} & 2.9723 \\
 & Glossy & 0.0661 & 0.0665 & \textbf{0.0463} &  & 0.2110 & \textbf{0.0675} & 0.0703 &  & 0.1124 & \textbf{0.0088} & 0.0129 &  & 39.7178 & \textbf{1.2403} & 2.9641 \\*[.5em]
Spiky sphere & Diffuse & 0.0667 & 0.0699 & \textbf{0.0606} &  & 0.1478 & 0.4398 & \textbf{0.0704} &  & 0.0783 & 0.1030 & \textbf{0.0130} &  & 35.0527 & 46.9025 & \textbf{2.9726} \\
 & Rough plastic & 0.0667 & 0.0661 & \textbf{0.0599} &  & 0.0961 & \textbf{0.0703} & 0.0705 &  & 0.0378 & 0.0157 & \textbf{0.0129} &  & 14.3937 & 4.4997 & \textbf{2.9793} \\
 & Glossy & 0.0668 & 0.0646 & \textbf{0.0614} &  & 0.1373 & \textbf{0.0649} & 0.0704 &  & 0.0623 & \textbf{0.0099} & 0.0128 &  & 22.3560 & \textbf{1.6903} & 2.9659 \\
 \bottomrule
\end{tabular}%
}
\vspace{.25em}
\caption{We evaluate our method on a synthetic dataset composed of 5 objects, each having 3 distinct materials, and compare it to two competing techniques: RANI~\cite{lombardi2016reflectance} and SH~\cite{ramamoorthi2001signal}. Our approach outperforms the others in most scenarios. In some instances, having the full transport matrix when the object is highly reflective (glossy) helps SH in recovering a more accurate estimate.}
\label{tab:quantitative_synthetic}
\end{table*}

\subsection{Lighting estimation from real data}
\label{sec:lighting_estimation_real}

\begin{figure}[t]
\centering
\includegraphics[width=0.8\linewidth]{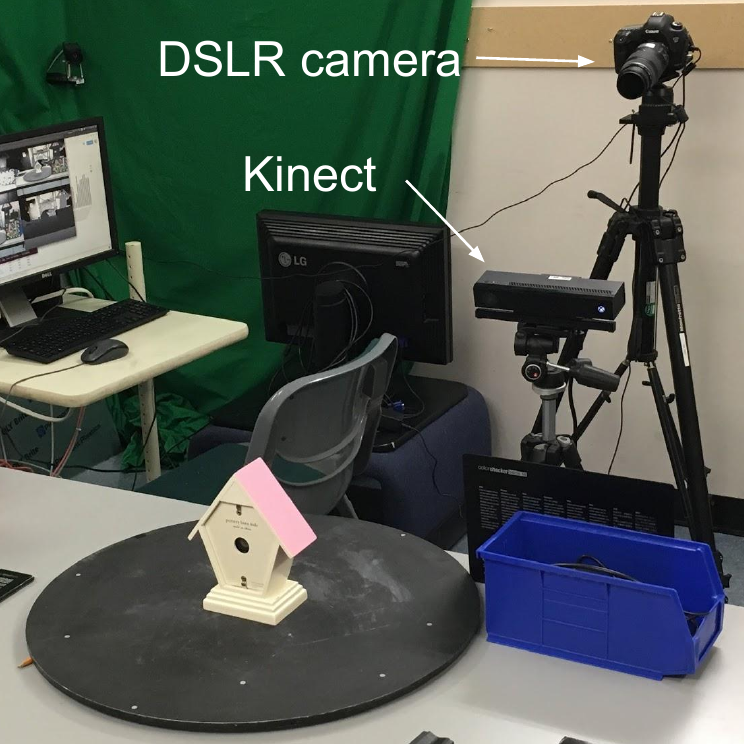}
\caption{Setup for real data capture. The Kinect V2 camera acquires the image of the object (here, a toy house), while the DSLR camera captures an HDR image of a mirror sphere (placed at the same position as the object) which will later be used as the GT environment map for that setup.}
\label{fig:data_capture}
\end{figure}

We perform experiments on real images of an object with known geometry. To acquire the data, we first obtain a detailed textured model of the object with a Creaform GoScan\texttrademark~sensor. We then place the object in front of a Kinect, as illustrated in fig.~\ref{fig:data_capture}, and obtain its pose with the 6-DOF RGBD object tracker from \cite{garon-tvcg-17}. The RGB frame from the Kinect is first linearized with a Macbeth chart, then used as an input to our network. We also capture the ground truth HDR lighting conditions for the scene by inserting a metallic sphere~\cite{Debevec-siggraph-98} and capturing a bracketed image sequence of 7 exposures with a Canon 5D Mark iii camera to ensure the entire dynamic range is properly captured, placed just above the Kinect. The exposures are then fused to a single HDR environment map, which is subsequently rotated manually to match the appearance of the sphere in the Kinect reference frame and re-exposed using the Macbeth chart to match the Kinect exposure. 

\begin{figure*}
\centering
\includegraphics[width=1\linewidth]{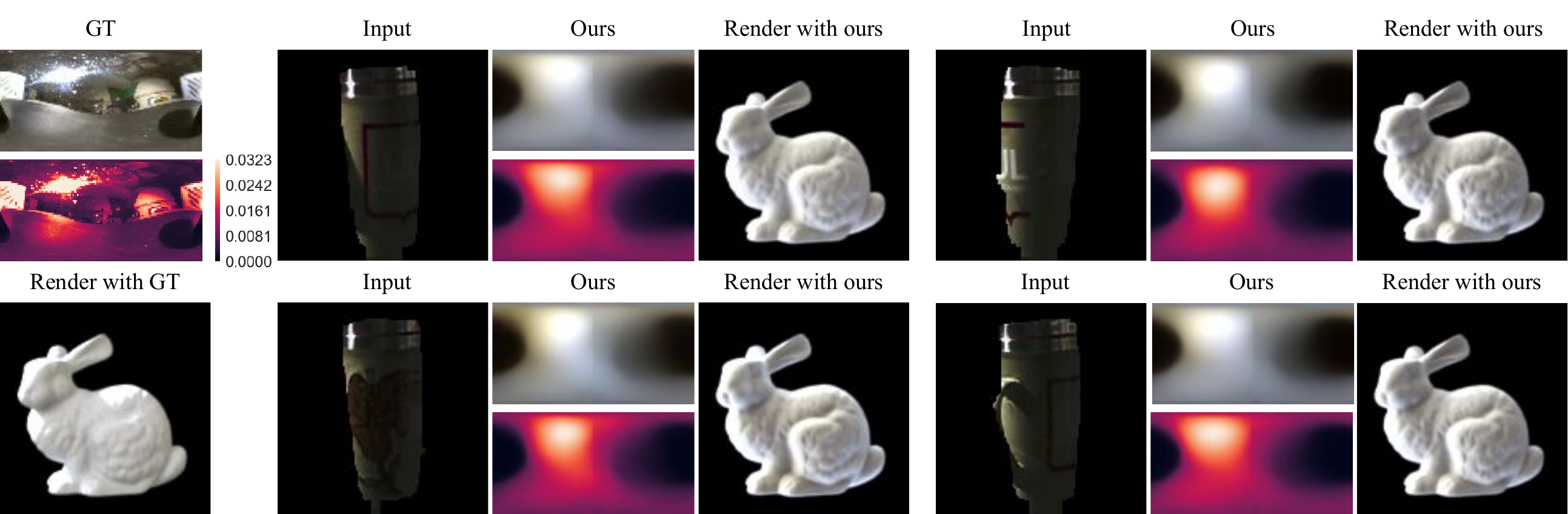}
\caption{Qualitative results on real data. One can see that, as the object turns, the lighting estimates provided by our method remains approximately constant.}
\label{fig:qualitative_illum_predict_real}
\end{figure*}

Next, with the pose provided by the tracker, we render the virtual version of the object to acquire the normal map and a mask of the object that will be used to remove the background. However, since tracking is not always perfectly accurate,the mask is eroded slightly to avoid having background pixels wrongly assigned to the object. Finally, to train the network, we create a synthetic LDR dataset with renders of the target object modeled with material properties that resemble the real object. The results for a cup can be seen in fig.~\ref{fig:qualitative_illum_predict_real}. We see that the network is capable of correctly predict the strongest light source even when the object undergoes rotation. The renderings shown below demonstrate that the recovered lighting conditions can be used to render an object in a way that is very similar to the ground truth (shown on the left). 

\section{Discussion}

In this paper, we have proposed a new learning method to efficiently estimate indoor lighting from a single picture of a known object. By leveraging a novel dataset of HDR panoramas with a convolutional autoencoder, we show that it is possible to 1) learn a compressed latent space for indoor lighting, and 2) to learn to map the appearance of the object to that space. Our experiments demonstrate that our lighting autoencoder can more compactly represent HDR environment maps than the ubiquitous spherical harmonics representation. Furthermore, we also show that our learning-based approach can infer lighting from an object more robustly than inverting the light transport matrix in a wide variety of scenarios. Finally, we demonstrate the applicability of our approach on real data. 


The main limitation of the proposed technique is that different illumination prediction networks must be re-trained for each object and each material properties. This has however limited negative impact in a practical scenario, where training can be done off-line once. However, another limitation is that the geometry and material properties of the object must be estimated to train the neural network. While it is relatively easy to obtain the geometry of a real object, it is not trivial to estimate its reflectance properties. For this reason, we believe that being able to estimate light regardless of the material is a promising future direction of research which we aim to investigate. Another limitation is that the network has difficulty in dealing with light sources of different colors. For example, the lighting environment in the middle-left example in fig.~\ref{fig:qualitative_illum_predict} contains roughly two light sources: a pink one on the left and a greenish one on the far right, which are both visible in the image of the reflective dragon. While SH is able to identify those colors, our neural network predicts instead two light sources of similar colors. We suspect the neural network fails to correctly identify the colors because this situation does not happen all that often in its training dataset. Despite these limitations, we believe that our work is a significant step towards a more realistic light estimation that can be readily be used for real-world applications like augmented reality. Integrating our lighting estimation with real-time tracking would be an interesting area of future research. 

\section*{Acknowledgements}

The authors thank S\'ebastien Poitras, Jean-Michel Fortin and Gabriel Lavin-Muller for their help in labeling the HDR dataset. This work was partially supported by an INO Ph.D. fellowship to Henrique Weber, the NSERC Discovery Grant RGPIN-2014-05314, and the REPARTI Strategic Network. We gratefully acknowledge the support of Nvidia with the donation of the GPUs used for this research. 

{\small
\bibliographystyle{ieee}
\bibliography{egbib}
}

\end{document}